\documentclass[journal]{IEEEtran}

\ifCLASSINFOpdf
\else
   \usepackage[dvips]{graphicx}
\fi
\usepackage{url}

\usepackage{graphicx}
\usepackage{cite}
\usepackage{subfigure}
\usepackage{amssymb}
\usepackage{booktabs}
\usepackage{multirow}
\usepackage{amsmath}
\usepackage{pifont}

\begin{document}

\title{AFNet-M: Adaptive Fusion Network with Masks for 2D+3D Facial Expression Recognition}

\author{Mingzhe~Sui, 	
	Hanting~Li,
	Zhaoqing~Zhu,
	and~Feng~Zhao
	\thanks{\textit{(Corresponding author: Feng Zhao.)}}
	\thanks{The authors are with the Department of Automation, School of Information Science and Technology, University of Science and Technology of China, Hefei 230027, China, and also with the National Engineering Laboratory for Brain-inspired Intelligence, Hefei 230027, China (e-mail: sa20@mail.ustc.edu.cn;  ab828658@mail.ustc.edu.cn;
	zhaoqingzhu@mail.ustc.edu.cn;
	fzhao956@ustc.edu.cn).}
}

\markboth{}
{Shell \MakeLowercase{\textit{et al.}}: Bare Demo of IEEEtran.cls for IEEE Journals}
\maketitle

\begin{abstract}
	
2D+3D facial expression recognition (FER) can effectively cope with illumination changes and pose variations by  simultaneously merging 2D texture and more robust 3D depth information. Most deep learning-based approaches employ the simple fusion strategy that concatenates the multimodal features directly after fully-connected layers, without considering the different degrees of significance for each modality. Meanwhile, how to focus on both 2D and 3D local features in salient regions is still a great challenge. In this letter, we propose the adaptive fusion network with masks (AFNet-M) for 2D+3D FER. To enhance 2D and 3D local features, we take the masks annotating salient regions of the face as prior knowledge and design the mask attention module (MA) which can automatically learn two modulation vectors to adjust the feature maps. Moreover, we introduce a novel fusion strategy that can perform adaptive fusion at convolutional layers through the designed importance weights computing module (IWC). Experimental results demonstrate that our AFNet-M achieves the state-of-the-art performance on BU-3DFE and Bosphorus datasets and requires fewer parameters in comparison with other models. 

\end{abstract}


\IEEEpeerreviewmaketitle

\section{Introduction}

\IEEEPARstart{F}{acial} expression is a significant means of nonverbal communication since it can express human cognition and emotions. In general, facial expression recognition (FER) aims to help machines infer six basic expressions, which are anger, disgust, fear, happiness, sadness, and surprise, and figures prominently in human-computer interaction areas~\cite{Li2020Deep, Guo2020Real, Lian2020Expression}. Although previous studies based on hand-crafted features~\cite{Happy2012A, Shi2020An} or deep learning~\cite{Wang2020Suppressing, Xia2021Local} have achieved excellent performance in 2D FER, it is under the premise of good image quality. The drastic variations in illumination and poses can still have a great impact on 2D FER~\cite{Li2020Deep}.

3D scans containing depth information perform better robustness to illumination and pose changes, and can also capture subtle muscle deformations~\cite{Li2020Deep}. Therefore, complementary multimodal 2D+3D FER has gradually attracted increasing attention in recent years. Li~\textit{et al}.~\cite{Li2017Multimodal} first introduced CNN to 2D+3D FER, where they represented each 3D scan as six attribute maps, and fed them into a deep fusion CNN with six branches for classification. Benefiting from the powerful learning ability of networks, the accuracy of deep learning-based models~\cite{Li2017Multimodal, Wei2018Unsupervised, Jan2018Accurate, Chen2018Fast, Tian20193D, Jiao2019Facial, Zhu2019Discriminative, jiao20202d+, Zhu2020Intensity, Sui2021FFNet, ni2022facial} has comprehensively surpassed the methods based on hand-crafted features~\cite{Tang20083D, Gong2009Automatic, Li20123D, Yang2015Automatic, Li2015An, Fu2019FERLrTc, Mpiperis2008Bilinear, Zhen2016Muscular, Berretti2010A, Lemaire2013Fully}. 

However, there still exist two issues in 2D+3D FER. First, most algorithms do not take good advantage of local features in salient regions (e.g., the neighborhoods of the mouth , nose, and eyes). Jiao~\textit{et al}.~\cite{Jiao2019Facial} proposed the FA-CNN to localize the discriminative facial parts, while the receptive fields will also focus on irrelevant areas such as the forehead, and the distribution is not stable enough from their visualization of heat maps. Sui~\textit{et al}.~\cite{Sui2021FFNet} designed the masks to directly enhance the local features in the whole salient regions, however, diverse components make various contributions to the judgment of one expression. For example, the features of the eyes and mouth are more critical than those of the nose. Thus, learning the distribution of salient regions discriminately from the masks is necessary. Another is that many deep models~\cite{Jiao2019Facial, Zhu2019Discriminative, Sui2021FFNet} employ the simple fusion strategy, which concatenates the multimodal features directly and equally after fully-connected layers. Nonetheless, at this time, the resolution of the features with each modality is too low and the ability to perceive local geometric details is poor, which is not conducive to the free attention flow among modalities~\cite{nagrani2021attention}. Furthermore, each modality places a different emphasis on the current classification task. We need to consider it before fusion.

To address the above problems, we propose the adaptive fusion network with masks (AFNet-M) for 2D+3D FER. The contributions of our model are as follows:

\begin{itemize}
	\item We design the mask attention module (MA) in the first half of out AFNet-M, which can automatically learn two modulation vectors from the masks annotating salient regions to enhance local features discriminatively. 
	\item We introduce a new fusion strategy that incorporates depth features into texture features in the second half of our model. Considering that the contribution rates of the features with two modalities to the results are different, we perform adaptive fusion by the designed importance weights computing module (IWC).
	\item Our AFNet-M achieves the highest accuracy on BU-3DFE and Bosphorus datasets and also demands fewer parameters in comparison with the state-of-the-art methods in 2D+3D FER.
\end{itemize}

\begin{figure*}[t]
	\begin{center}
		\includegraphics[width=0.8\linewidth]{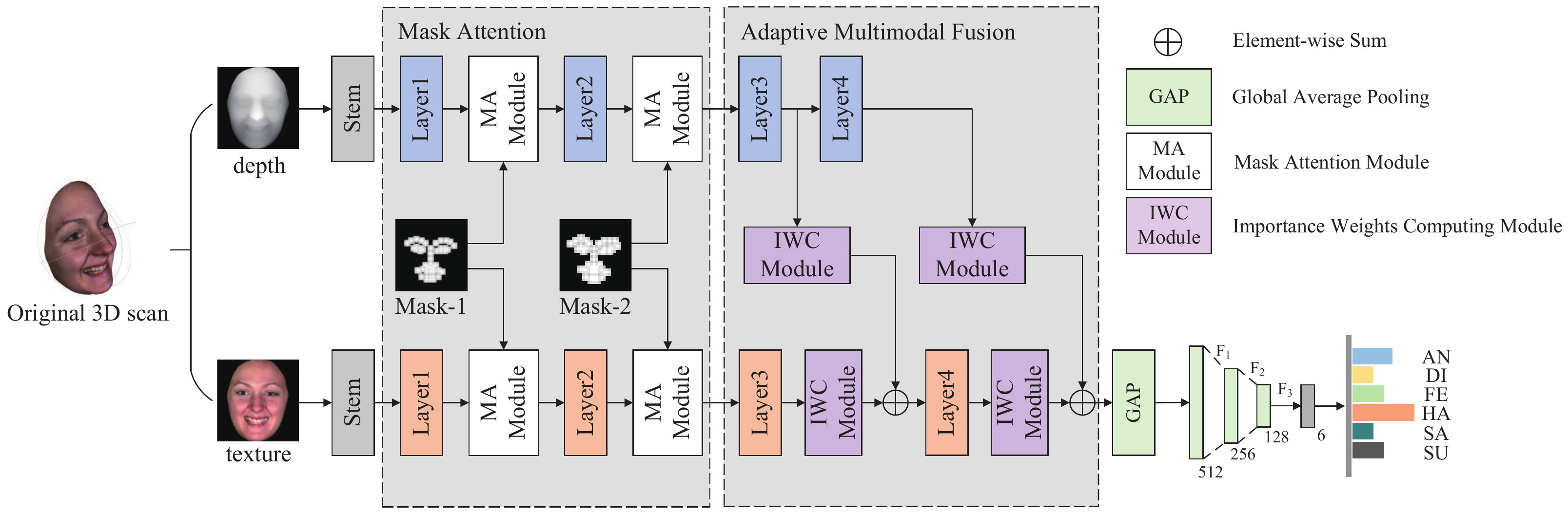}
		\caption{The pipeline of our AFNet-M. STEM means all the operations before the first residual block in ResNet18, including a $7 \times 7$ convolutional layer and a max pooling layer. The softmax function is utilized to predict the expression.}
		\label{figure1}
		\vspace{-0.4cm}	
	\end{center}
\end{figure*}

\begin{figure}[t]
	\begin{center}
		\includegraphics[width=1.0\linewidth]{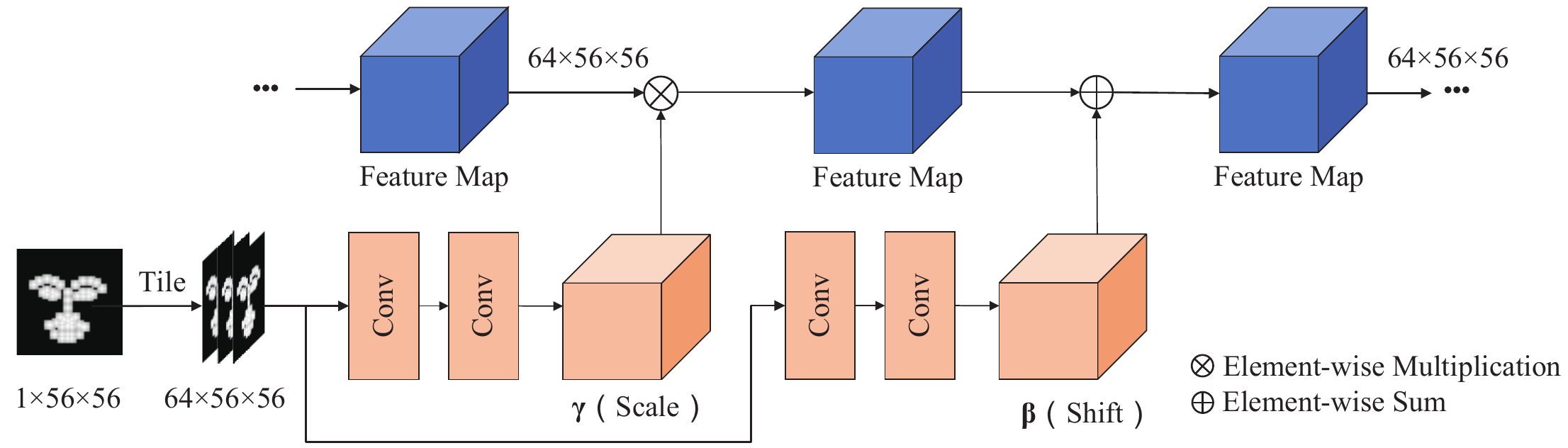}
		\caption{The structure of the proposed MA module at Layer1.}
		\label{figure2}
		\vspace{-0.4cm}	
	\end{center}	
\end{figure}

\section{METHODOLOGIES}

\subsection{Overview of AFNet-M}
The framework of our AFNet-M is illustrated in Fig.~\ref{figure1}. The whole network is based on dual-branch ResNet18s~\cite{He2016Deep}. In preprocessing, we execute gridfit algorithm~\cite{S2018Learning} to generate aligned texture and depth images from each 3D scan and perform surface processing~\cite{Sui2021FFNet} containing outlier removal, hole filling, and noise removal to improve their quality. We also generate the masks annotating salient regions at two scales (Mask-1 is 56$\times$56 and Mask-2 is 28$\times$28) as~\cite{Sui2021FFNet} did. In feature extraction, we first design the MA module at Layer1 and Layer2 and introduce the generated masks as prior knowledge to enhance depth and texture local features separately in the spatial dimension. Next, at Layer3 and Layer4, we use the devised IWC module to perform the adaptive fusion of depth and texture features with weight coefficients. Finally, the multimodal representations are used for classification after three fully-connected layers and the softmax function.

\subsection{Mask Attention}
Different from directly enhancing all salient regions~\cite{Sui2021FFNet}, we consider learning the distribution of salient regions discriminately from the masks and propose the MA module, as shown in Fig.~\ref{figure2}. The MA module consists of four convolutions. Taking the texture branch as an example, we first reshape the single-channel mask to 64$\times$56$\times$56 at Layer1 to match the texture feature maps. Two modulation vectors ($\boldsymbol{\gamma}$ and $\boldsymbol{\beta}$) are automatically learned separately through two independent convolution groups, which represent the distribution of salient regions in the masks. To reduce extra parameters, the kernel size of all convolutions is 1$\times$1. The process of enhancing the texture feature maps can be expressed as:
\begin{equation}
\tilde{\mathbf{X}_t} = \boldsymbol{\gamma} \otimes \mathbf{X}_t + \boldsymbol{\beta}
\end{equation}
where $\tilde{\mathbf{X}_t}$ and $\mathbf{X}_t$ represent the texture feature maps after and before enhancement, respectively. $\otimes$ represents the element-wise multiplication. The shapes of $\boldsymbol{\gamma}$ and $\boldsymbol{\beta}$ are the same as $\mathbf{X}_t$. It is equivalent that we scale and shift the feature map of each channel of $\mathbf{X}_t$ in the spatial dimension. We do the same process for the depth images. During the training stage, the two modulation vectors will be continuously adjusted to make the network discriminately enhance the depth and texture local features in salient regions of the face. We do not use the MA module in the second half in that the landmarks cannot be detected to generate the mask and the the receptive field of each pixel has almost covered the entire input image when the spatial size is too small.

\subsection{Adaptive Multimodal Fusion}
\begin{figure}[t]
	\begin{center}
		\includegraphics[width=1.0\linewidth]{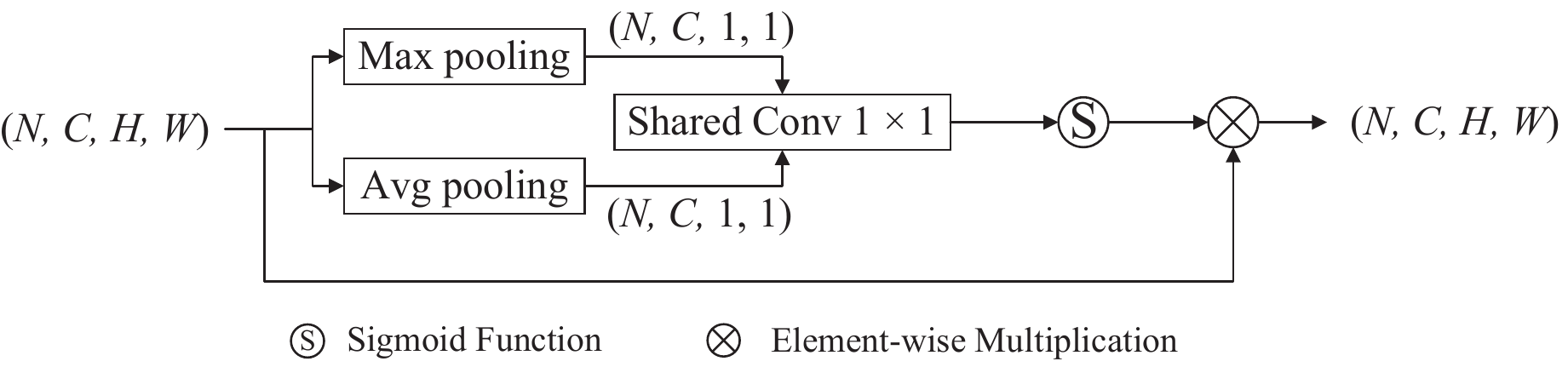}
		\caption{The structure of the proposed IWC module. $N$, $C$, $H$, and $W$ denote the size of the batch, channels, height and width, respectively.}
		\label{figure3}
		\vspace{-0.4cm}	
	\end{center}	
\end{figure}

The interaction among multimodal features is the key to multimodal fusion tasks. To fully utilize the multiscale features at convolutional layers and form more comprehensive multimodal representations, we incorporate depth features into texture features at Layer3 and Layer4, as shown in Fig.~\ref{figure2}. Considering that the contribution rates of the features with two modalities to the classification are different, inspired by ACM block~\cite{Hu2019ACNET}, we introduce the IWC module to perform adaptive fusion, as depicted in Fig.~\ref{figure3}. We compute an importance weight for each channel of the feature maps, formulated as:
\begin{equation}
	\resizebox{0.90\hsize}{!}{$\begin{aligned}
	\mathbf{t}_{iw}=Sigmoid(Conv(AvgPool(\mathbf{X}_t))+Conv(MaxPool(\mathbf{X}_t)))
	\end{aligned}$}	
\end{equation}
where $\mathbf{X}_t$ is the texture feature maps at Layer3 or Layer4, $\mathbf{t}_{iw}$ ranging from 0-1 represents the texture importance weights. $Conv$ represents the shared convolution with the kernel size of 1$\times$1, which can mine the correlations among channels. Similarly, we also compute the depth importance weights $\mathbf{d}_{iw}$. The adaptive fusion can be expressed as:
\begin{equation}
\left\{\begin{array}{l}
\hat{\mathbf{X}_d} = \mathbf{d}_{iw} \otimes \mathbf{X}_d
\vspace{0.1cm}\\
\hat{\mathbf{X}_t} = \mathbf{t}_{iw} \otimes \mathbf{X}_t
\vspace{0.1cm}\\
\mathbf{M} = \hat{\mathbf{X}_t} + \hat{\mathbf{X}_d}
\end{array}\right.
\end{equation}
where $\mathbf{M}$ represents the formed multimodal features after adaptive fusion. Therefore, the channel features with some modality that are more crucial to the results will also account for a larger proportion of the formed multimodal representations, while those that have a negative impact on the classification will be suppressed to a certain extent. The reason why we choose to perform adaptive fusion at Layer3 and Layer4 is that the resolution and the ability to perceive local geometric details of the features with each modality are appropriate. We also give the ablation studies to prove this in section \uppercase\expandafter{\romannumeral3}.

\section{Experimental Results}

\subsection{Datasets and Evaluation Protocol}

\textbf{BU-3DFE dataset}. The BU-3DFE dataset~\cite{Yin2006A} comprises 100 subjects with ages from 18 to 70. Each subject contains six prototypical expressions (i.e., anger, disgust, fear, happiness, sadness, and surprise), which are elicited by various manners with four levels of expression intensity. 

\textbf{Bosphorus dataset}. The Bosphorus dataset~\cite{Savran2008Bosphorus} consists of 4666 3D scans from 105 subjects with ages from 25 to 35. Different from BU-3DFE dataset, only 63 subjects contain six prototypical expressions with one level of expression intensity. 

\textbf{Evaluation Protocol}. To compare equally with other methods, we follow the standard protocol applied in~\cite{Li2017Multimodal, Zhu2019Discriminative, Jiao2019Facial, Zhu2020Intensity, jiao20202d+, Sui2021FFNet, ni2022facial} to evaluate our AFNet-M. In this protocol, 60 subjects with level 3 and level 4 expression intensity from 100 subjects in BU-3DFE dataset and 60 subjects from 63 subjects in Bosphorus dataset are selected randomly, which are fixed in the whole experiments. Then, the average accuracy of 100 times of 10-fold cross-validation is executed to evaluate the model for more stable and reliable results. 

\subsection{Implementation Details}

The depth and texture images are resized to 3$\times$224$\times$224 after preprocessing. At the training stage, we initialize two ResNet18s in the AFNet-M with the pre-trained parameters on ImageNet~\cite{Deng2009Imagenet}. All the convolutions in the MA and IWC module follow a normal distribution with the mean of 0 and the standard deviation of 0.02, and the bias is set to 0. The Adam optimizer with betas $(0.9, 0.999)$ is adopted. In addition, we set the learning rate to 0.0001 and fix it in 70 epochs for the cross-entropy loss function. All the experiments are conducted on one NVIDIA GeForce RTX3070 card with Pytorch.

\subsection{Results}

\textbf{Comparisons with the state-of-the-art methods}. Table~\ref{table1} shows the performance comparisons of our model with other approaches on BU-3DFE and Bosphorus. We can see that our AFNet-M outperforms state-of-the-art methods with the highest accuracy of 90.08\% and 88.31\%, whether compared with hand-crafted features or deep networks. Moreover, compared with the results under single modality input, merging 2D and 3D features can significantly boost the accuracy. 

\begin{figure}[t]
	\begin{center}			
		\subfigure[\scriptsize{Ours on BU-3DFE}]{
			\label{CM_AFNet-M_BU3DFE}
			\begin{minipage}[b]{0.4\linewidth}
				\includegraphics[width=1.0\linewidth]{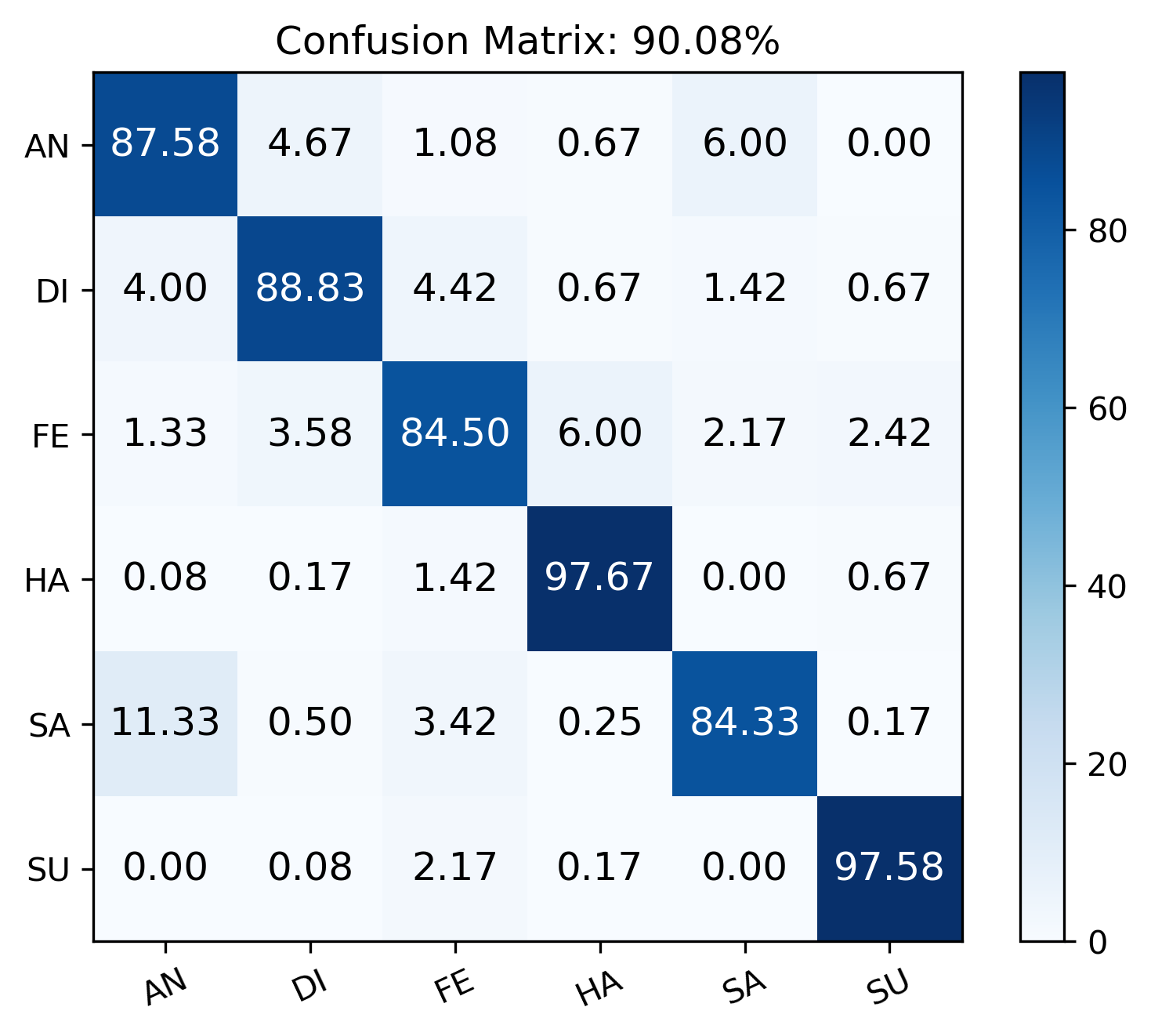}
			\end{minipage}}	
		\subfigure[\scriptsize{Ours on Bosphorus}]{
			\label{CM_AFNet-M_Bosphorus}
			\begin{minipage}[b]{0.4\linewidth}
				\includegraphics[width=1.0\linewidth]{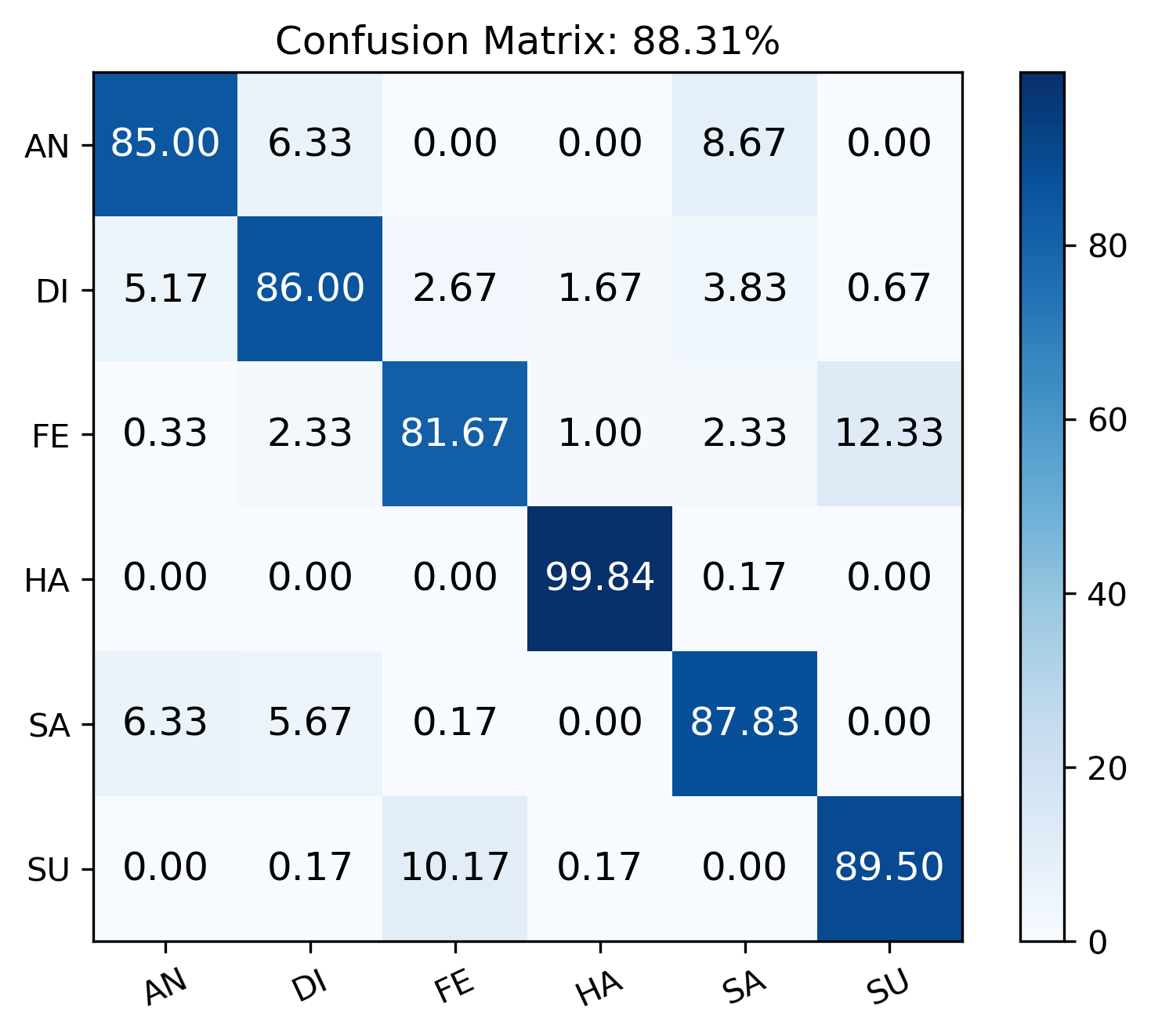}
			\end{minipage}}	
		\subfigure[\scriptsize{Ours w/o MA on BU-3DFE}]{
			\label{CM_AFNet_BU3DFE}
			\begin{minipage}[b]{0.4\linewidth}
				\includegraphics[width=1.0\linewidth]{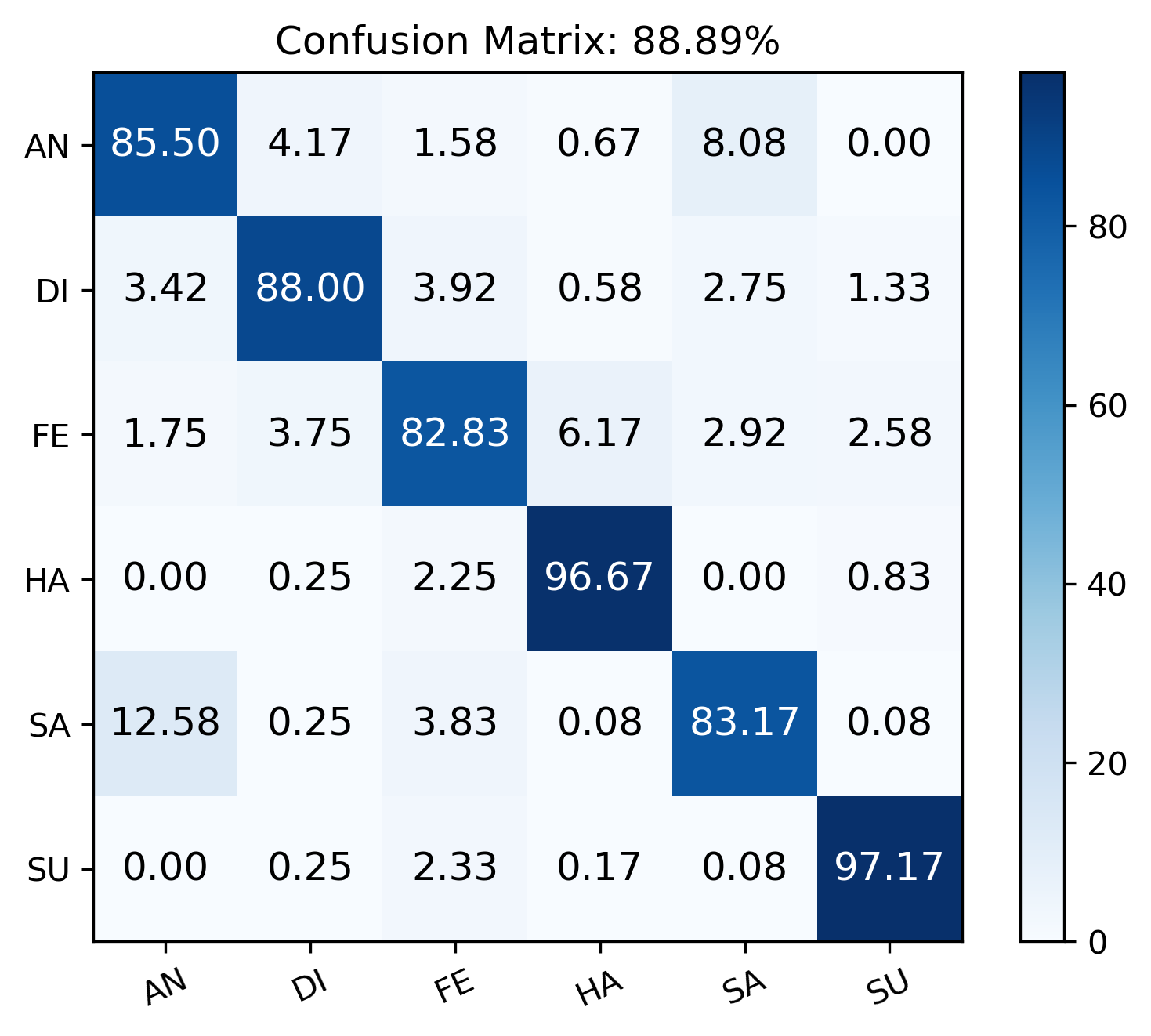}
			\end{minipage}}	
		\subfigure[\scriptsize{Ours w/o MA on Bosphorus}]{
		\label{CM_AFNet_Bosphorus}
			\begin{minipage}[b]{0.4\linewidth}
			\includegraphics[width=1.0\linewidth]{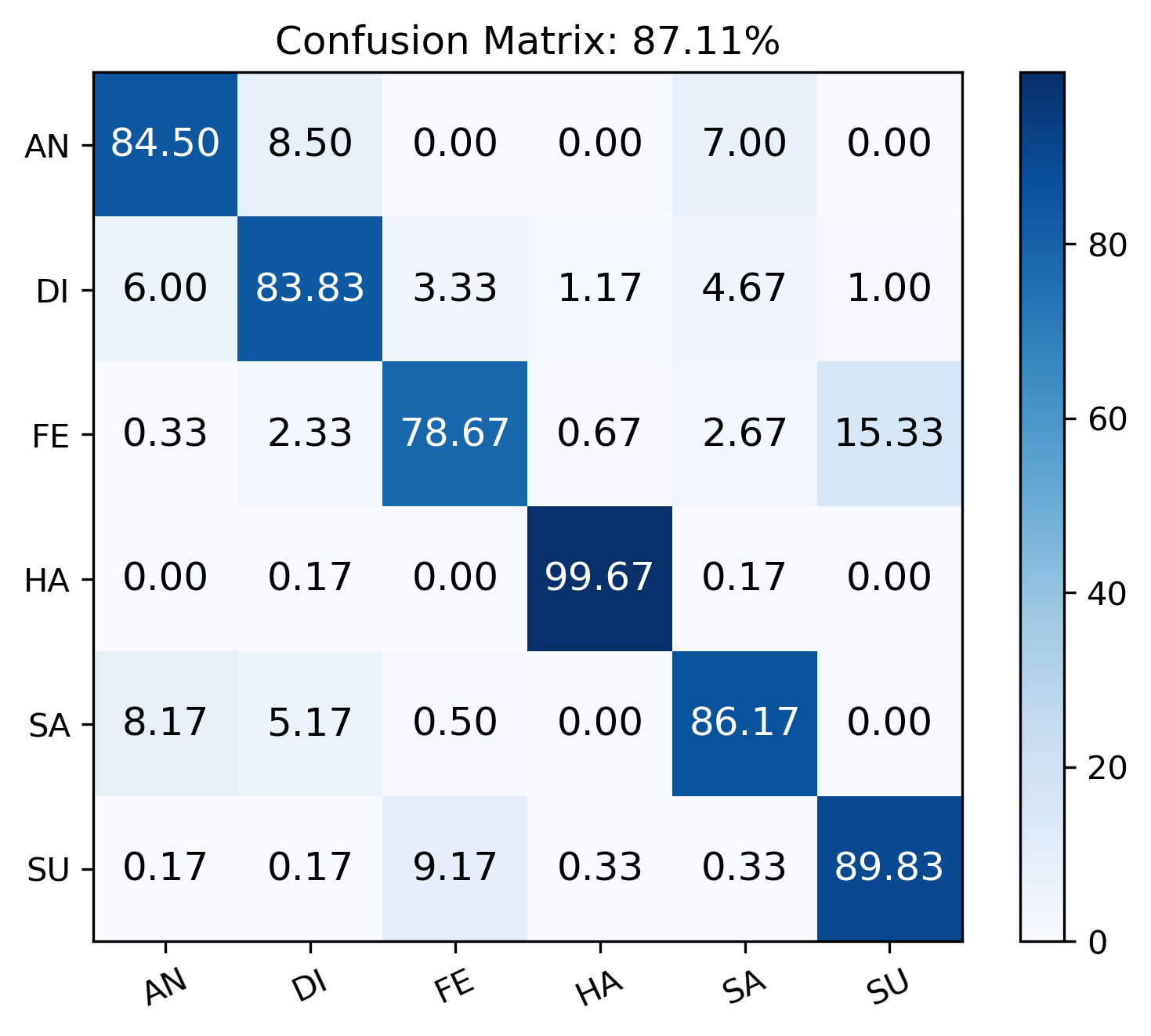}
			\end{minipage}}	
		\caption{Confusion matrices on BU-3DFE and Bosphorus. AN, DI, FE, HA, SA, and SU represent anger, disgust, fear, happiness, sadness, and surprise expression, respectively.}	
		\label{figure4}	
		\vspace{-0.4cm}	
	\end{center}
\end{figure}

\textbf{Confusion matrix}. To reflect the performance for each category, we also give the confusion matrices, as shown in Fig.~\ref{CM_AFNet-M_BU3DFE} and (b). We can find that happiness and surprise have better results, while the accuracies for the remaining are average, which is because that the features of the former with exaggerated muscle deformations are more discriminate, and the latter are easily confused with each other.

\begin{table}[t]\footnotesize
	\renewcommand\arraystretch{1.0}
	\begin{center}
		\caption{Comparison results on BU-3DFE and Bosphorus. HC and DL represent hand-crafted and deep learning-based features.} \label{table1}
		\begin{tabular*}{\hsize}{cccccc}
			\toprule  
			Method & Year & Data & Feature & BU-3DFE & Bosphorus \\
			\midrule
			Tang~\textit{et al}.~\cite{Tang20083D} & 2008 & 3D & HC & 74.51 & - \\
			Gong~\textit{et al}.~\cite{Gong2009Automatic} & 2009 & 3D & HC & 76.22 & - \\
			Li~\textit{et al}.~\cite{Li20123D} & 2012 & 3D & HC & 80.14 & 75.83 \\
			Yang~\textit{et al}.~\cite{Yang2015Automatic} & 2015 & 3D & HC & 84.80 & 77.50 \\
			Li~\textit{et al}.~\cite{Li2015An} & 2015 & 2D+3D & HC & 86.32 & 79.72 \\
			Fu~\textit{et al}.~\cite{Fu2019FERLrTc} & 2019 & 2D+3D & HC & 82.89 & 75.93 \\
			\midrule
			Li~\textit{et al}.~\cite{Li2017Multimodal} & 2017 & 2D+3D & DL & 86.86 & 80.28 \\
			Wei~\textit{et al}.~\cite{Wei2018Unsupervised} & 2018 & 2D+3D & DL & 88.03 & 82.50 \\
			Jan~\textit{et al}.~\cite{Jan2018Accurate} & 2018 & 2D+3D & DL & 88.54 & - \\
			Chen~\textit{et al}.~\cite{Chen2018Fast} & 2018 & 3D & DL & 86.67 & - \\
			Tian~\textit{et al}.~\cite{Tian20193D} & 2019 & 2D+3D & DL & - & 79.17 \\
			Jiao~\textit{et al}.~\cite{Jiao2019Facial} & 2019 & 2D+3D & DL & 89.11 & - \\
			Zhu~\textit{et al}.~\cite{Zhu2019Discriminative} & 2019 & 2D+3D & DL & 88.35 & - \\
			Jiao~\textit{et al}.~\cite{jiao20202d+} & 2020 & 2D+3D & DL & 89.72 & 83.63 \\
			Zhu~\textit{et al}.~\cite{			Zhu2020Intensity} & 2020 & 2D+3D & DL & 88.75 & - \\
			Sui~\textit{et al}.~\cite{Sui2021FFNet} & 2021 & 2D+3D & DL & 89.82 & 87.65 \\
			Ni~\textit{et al}.~\cite{ni2022facial} & 2022 & 2D+3D & DL & 88.91 & 85.16 \\
			\midrule
			Ours & - & 2D & DL & 87.68 & 85.42 \\
			Ours & - & 3D & DL & 86.97 & 82.06 \\
			Ours & - & 2D+3D & DL & $\textbf{90.08}$ & $\textbf{88.31}$ \\
			\bottomrule
		\end{tabular*}
		\vspace{-0.4cm}	
	\end{center}
\end{table}

\begin{table}[t]\footnotesize
	\renewcommand\arraystretch{1.0}
	\begin{center}
		\caption{Ablation experiments of the fusion strategy.} \label{table2}
		\begin{tabular*}{\hsize}{ccccccc}
			\toprule  
			\multirow{2.5}{*}{Method} & \multicolumn{3}{c}{Fusion Strategy} & \multirow{2.5}{*}{IWC} & \multirow{2.5}{*}{BU-3DFE}& \multirow{2.5}{*}{Bosphorus} \\
			\cmidrule{2-4}
			& Data & Feature & Decision \\
			\midrule
			S1 & \checkmark & & &  & 86.54 & 85.08 \\
			S2 & & & \checkmark &  & 86.28 & 84.36 \\
			S3(FC) & & \checkmark & &  & 87.61 & 85.54 \\
			S4 & & \checkmark & & & 88.12 & 86.17 \\
			\midrule
			Ours & & \checkmark & & \checkmark & $\textbf{88.89}$ & $\textbf{87.11}$ \\
			\bottomrule
		\end{tabular*}
	\end{center}
	\vspace{-0.4cm}
\end{table}

\begin{figure}[t]
	\begin{center}			
		\subfigure[\scriptsize{Fusion positions on BU-3DFE}]{
			\label{position_BU3DFE}
			\begin{minipage}[b]{0.45\linewidth}
				\includegraphics[width=1.0\linewidth]{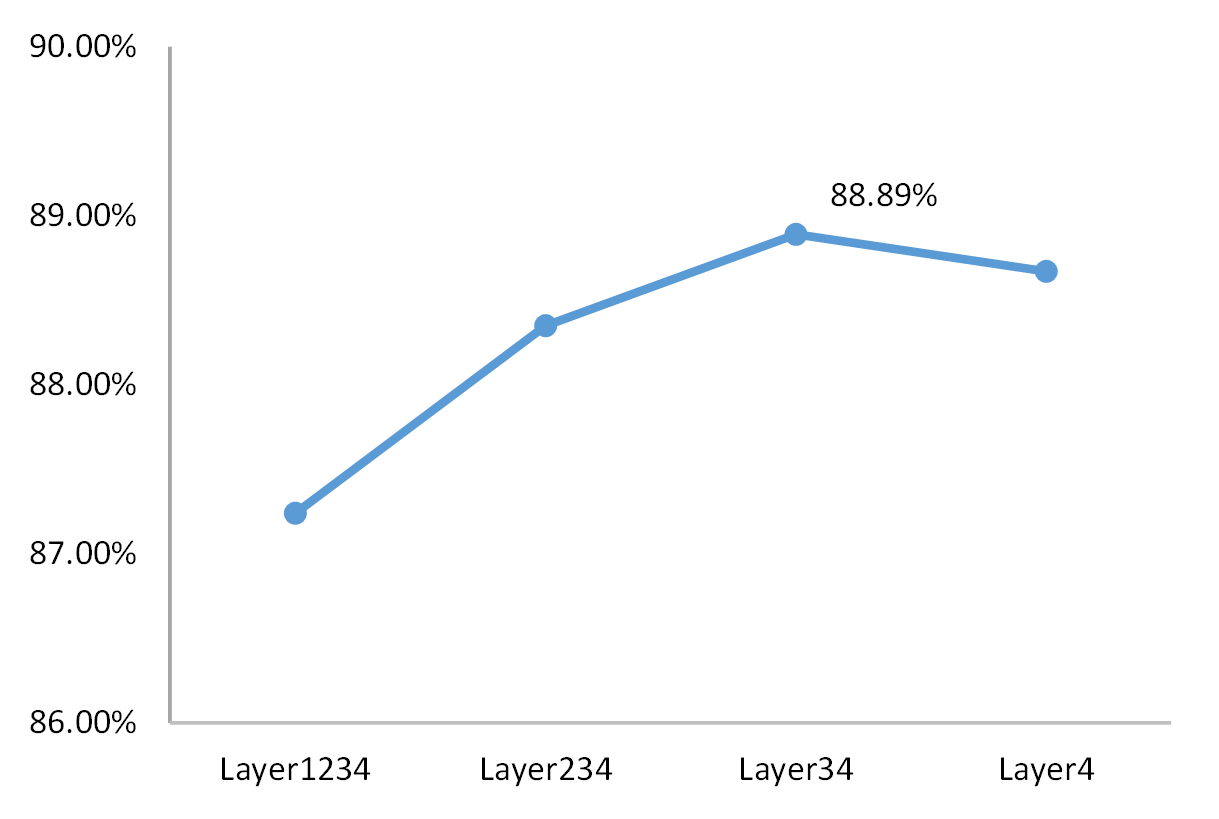}
		\end{minipage}}	
		\subfigure[\scriptsize{Fusion positions on Bosphorus}]{
			\label{position_Bosophorus}
			\begin{minipage}[b]{0.45\linewidth}
				\includegraphics[width=1.0\linewidth]{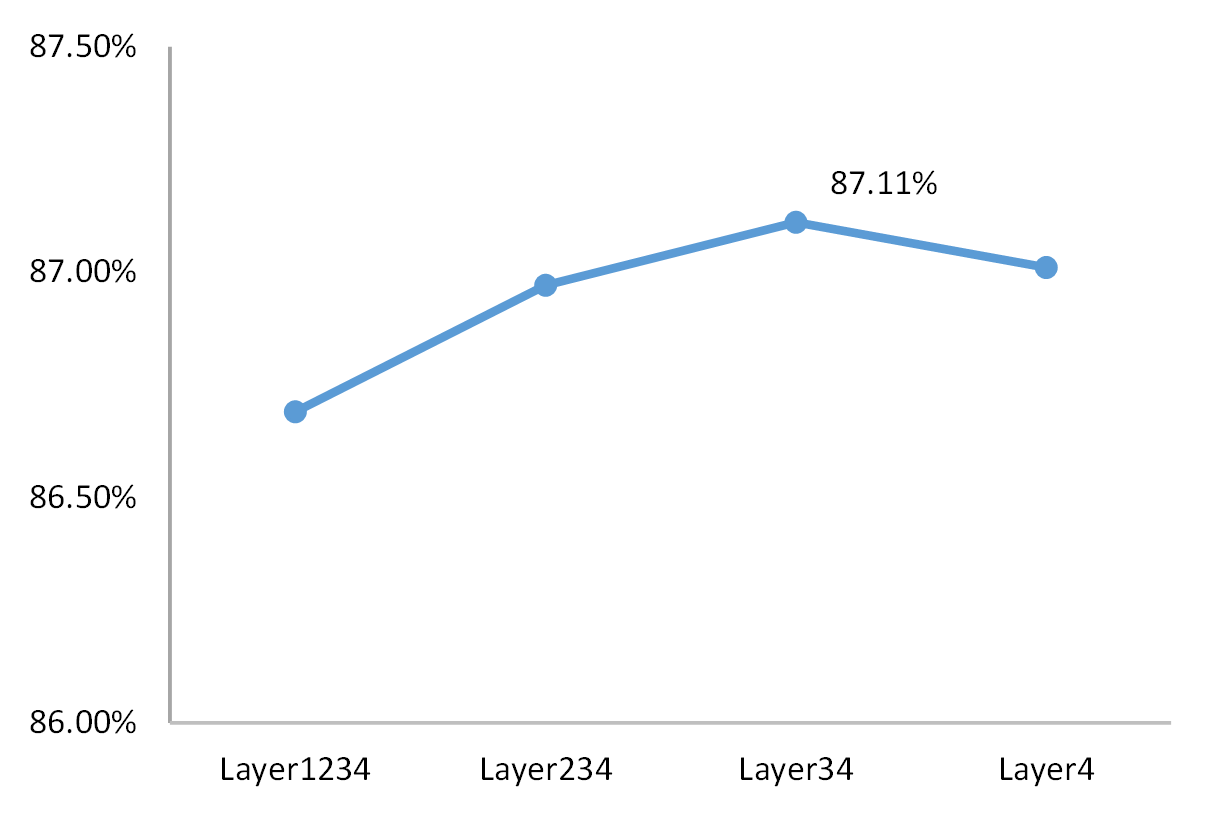}
		\end{minipage}}	
		\caption{Ablation experiments of choosing the fusion positions.}	
		\label{figure5}	
		\vspace{-0.4cm}	
	\end{center}
\end{figure}

\begin{table}[!htbp]\footnotesize
	\renewcommand\arraystretch{1.0}
	\begin{center}
		\caption{Ablation experiments of the MA module.} \label{table3}
		\begin{tabular}{ccccc}
			\toprule  
			Data & MA & Fusion & BU-3DFE & Bosphorus \\
			\midrule
			2D & & & 85.06 & 82.69 \\
			2D & \checkmark & & 87.68 & 85.42 \\
			3D & & & 84.37 & 80.58 \\
			3D & \checkmark & & 86.97 & 82.06 \\
			2D+3D & & \checkmark & 88.89 & 87.11 \\
			2D+3D & \checkmark & \checkmark & $\textbf{90.08}$ & $\textbf{88.31}$ \\
			\bottomrule
		\end{tabular}
	\end{center}
	\vspace{-0.4cm}
\end{table}

\subsection{Ablation Studies}
To prove the validity of each component in our AFNet-M, we conduct extensive ablation studies on two datasets.

\textbf{Evaluation of the fusion strategy}.
We evaluate different fusion strategies without the MA module, as shown in Table~\ref{table2}. S1, S2, and S3(FC) represent the fusion strategy at data level, decision level, and fully-connected layer at feature level. S4 is our proposed fusion strategy without the IWC module. Compared with the first three strategies, S4 achieves the highest accuracy, indicating that fusing the features of the convolutional layers can obtain multimodal features with better representations. From the last two rows in Table~\ref{table3}, we can see that the IWC module calculates the importance weights for the features with each modality to perform adaptive fusion, which can further improve the performance. Moreover, we also give the ablation experiments of choosing the fusion positions, as pictured in Fig.~\ref{figure5}. The results in both subfigures prove the correctness of choosing Layer3 and Layer4 for fusion. For Layer1234, it begins since the first residual block. The receptive field of the underlying convolution kernel is relatively small, too much edge information may be extracted, which is not conducive to multimodal fusion.

\textbf{Evaluation of the MA module}.
We evaluate the MA module with different modalities, as illustrated in Table~\ref{table3}. From the last two rows, we can see that using the proposed MA module to enhance local features discriminatively can improve the accuracy by 1.19\% and 1.2\% on BU-3DFE and Bosphorus under multimodal input. Furthermore, it can also boost the performance for a single 2D or 3D modality. From the comparisons of Fig. 4(c) and (a), (d) and (b), we can find that the masks annotating salient regions can help improve the recognition accuracies for almost all six categories, which fully shows the effectiveness of the designed MA module.

\begin{figure}[t]
	\begin{center}
		\includegraphics[width=0.8\linewidth]{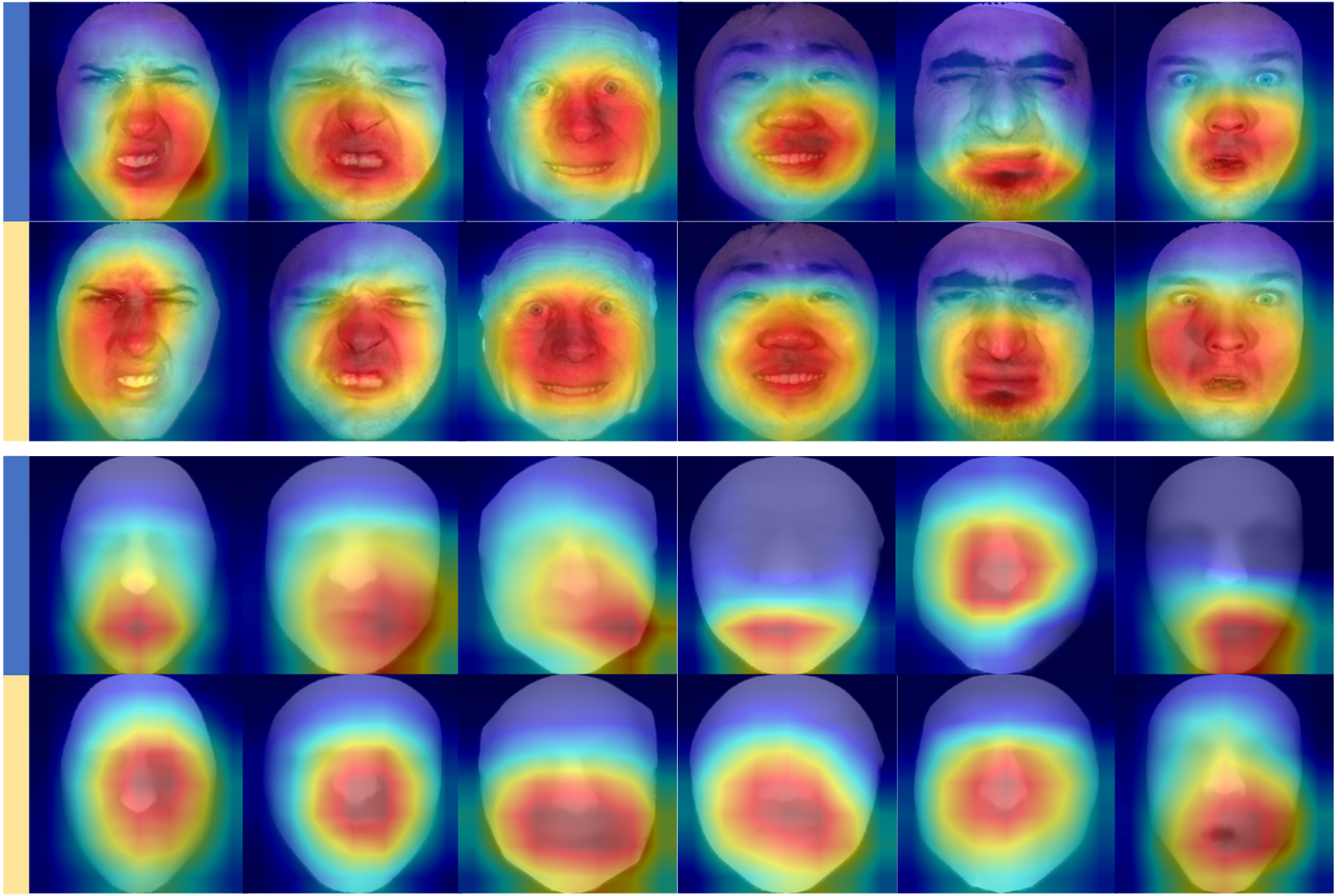}
		\caption{Visualization of heat maps with six basic expressions. The 1st and 2nd rows are texture heat maps w/o and with the MA module. The 3rd and 4th rows are depth heat maps w/o and with the MA module.}
		\label{figure6}
		\vspace{-0.4cm}	
	\end{center}	
\end{figure}

To reflect the distribution of the region of interest after employing the MA module, we also visualize the depth and texture heat maps with Grad-CAM~\cite{Selvaraju2016Grad}, as given in Fig.~\ref{figure6}. The 2nd and 4th rows obviously show that our AFNet-M has a more stable and concentrated region of interest for all expressions. More importantly, the masks enable the network to utilize the features of multiple facial parts (such as the eyes and mouth) to jointly determine the classification, rather than depending on a single part (see the anger expression in the 1st column in Fig.~\ref{figure6}).

\begin{table}[t]\footnotesize
	\renewcommand\arraystretch{1.0}
	\begin{center}
		\caption{Comparisons of the parameters} \label{table4}
		\begin{tabular}{ccccc}
			\toprule  
			Method & Year & Data & Parameters \\
			\midrule
			VGG-M-DF~\cite{Jan2018Accurate} & 2018 & 2D+3D & $\approx$ 327 \\
			DA-CNN~\cite{Zhu2019Discriminative} & 2019 & 2D+3D & $\approx$ 463\\
			FFNet-M~\cite{Sui2021FFNet} & 2021 & 2D+3D & $\approx$ 93 &  \\
			\midrule
			Ours w/o MA & - & 2D+3D & 90.51 \\
			Ours w/o IWC & - & 2D+3D & 86.53 \\
			AFNet-M & - &2D+3D & 91.54   \\
			\bottomrule
		\end{tabular}
	\end{center}
	\vspace{-0.4cm}
\end{table}

\subsection{Parameters Analysis}

To analyze the scale of the network, Table~\ref{table4} illustrates the comparison results of parameters. We can see that our AFNet-M achieves the highest accuracy with relatively minimal parameters (91.54 MB) compared with the state-of-the-art methods. And the incorporation of the designed MA module and IWC module only need tiny extra parameters.

\section{Conclusion}

In this letter, we propose a novel adaptive fusion network with masks (AFNet-M) for 2D+3D FER. Based on the generated masks annotating salient regions, we design the MA module which can automatically learn two modulation vectors to discriminatively enhance the 2D and 3D local features. To form better multimodal representations, we introduce a new fusion strategy that compute the importance weights for each modality through the devised IWC module to perform adaptive fusion. Extensive experimental results show that our AFNet-M has superior performance compared with the state-of-the-art methods on BU-3DFE and Bosphorus datasets, requiring fewer parameters and less memory costs simultaneously.

In the future, we will directly use the original 3D data and explore an end-to-end framework without the complicated preprocessing for multimodal 2D+3D FER.

\clearpage
\bibliographystyle{IEEEtran}
\bibliography{SPL_2022}

\end{document}